\documentclass{article}

\usepackage[nonatbib,preprint]{neurips_2022}

\usepackage[utf8]{inputenc} % allow utf-8 input
\usepackage[T1]{fontenc}    % use 8-bit T1 fonts
\usepackage{hyperref}       % hyperlinks
\usepackage{url}            % simple URL typesetting
\usepackage{booktabs}       % professional-quality tables
\usepackage{amsfonts}       % blackboard math symbols
\usepackage{nicefrac}       % compact symbols for 1/2, etc.
\usepackage{microtype}      % microtypography
\usepackage{xcolor}         % colors
\usepackage{amsmath, amssymb}
\usepackage{siunitx}
\usepackage{graphicx}

\title{Structural Dropout for Model Width Compression}

\author{
  Julian Knodt \\
  Princeton University\\
  Princeton, NJ 08544 \\
  \texttt{jknodt@princeton.edu} \\
}

\begin{document}

\maketitle

\begin{abstract}
  Existing ML models are known to be highly over-parametrized, and use significantly more resources than required for a given task. Prior work has explored compressing models offline, such as by distilling knowledge from larger models into much smaller ones. This is effective for compression, but does not give an empirical method for measuring how much the model can be compressed, and requires additional training for each compressed model. We propose a method that requires only a single training session for the original model and a set of compressed models. The proposed approach is a ``structural'' dropout that prunes all elements in the hidden state above a randomly chosen index, forcing the model to learn an importance ordering over its features. After learning this ordering, at inference time unimportant features can be pruned while retaining most accuracy, reducing parameter size significantly. In this work, we focus on Structural Dropout for fully-connected layers, but the concept can be applied to any kind of layer with unordered features, such as convolutional or attention layers. Structural Dropout requires no additional pruning/retraining, but requires additional validation for each possible hidden sizes. At inference time, a non-expert can select a memory versus accuracy trade-off that best suits their needs, across a wide range of highly compressed versus more accurate models.
\end{abstract}

%auto-ignore
\section{Introduction}

A neural network consists of some number of unstructured vector transformations from some input to some output. By learning parameters for these transformations, a neural net is able to optimize for a loss function, using gradient descent. While it has been shown that an infinitely large neural network can be a universal approximation for any function, it is not clear how \textit{small} neural networks can be, and how to efficiently find this size. With enough time and resources, it is possible to perform an exhaustive grid search to find hyper-parameters for model size, or use tools such as neural architecture search. For an ML researcher, this may be infeasible, and they will select hyper-parameters which are certainly too large, but definitely provide enough capacity to learn the task at hand. While this is practical for a researcher, for a user of these models they may be wasting computational resources, in memory and run-time efficiency. Furthermore, the burden is pushed onto the user to compress the model. To help compress models, there have been a large number of non-intrusive techniques for shrinking the size of models, such as quantization from 32 bit floating point numbers to 16 or even 8 bits, providing 2x and 4x memory savings respectively. There have also been a number of other techniques, such as knowledge distillation, where a much smaller student model is trained to match the output of a larger teacher model.

While some of these methods work out of the box on existing models, such as floating point quantization, other methods still leave choices such as the size of the compressed model, and may introduce additional error by retraining a new model. Thus, these methods may be non-starters for a user who wants to deploy a model, where a given model is too large for their resource budget. A better approach for model compression would allow for selecting from a resource usage to accuracy table provided alongside a compressible model, with no additional retraining. The reason that neural nets cannot simply be trimmed is because of the \textbf{unstructured} nature of the internal hidden state. That is, given some function $f(x) = g(h(x))$, where the target is $h^{*} = \text{argmin}_{h} g(h(x))$ and $h(x)$ is the hidden state of a network, and $h_2(x)\subset h_1(x)\subseteq h(x)$, it is not clear whether $g(h_2(x)) \stackrel{?}{<} g(h_1(x))$. It would be possible to evaluate this for all possible subsets, but this is computationally infeasible, since there are $2^{|h(x)|}$ possible sub-networks. If we can impose an ordering such that for some known choices of $h_1, h_2: h_2 \subset h_1 \implies g(h_2(x)) \stackrel{\text{likely}}{\leq} g(h_1(x))$, then we can prune the features $h_1\setminus h_2$, and measure the loss in performance without needing an exhaustive search. In addition, if we find that $g(h_2(x)) \geq g(h_1(x))$, we can even compress the network without accuracy loss.

We seek to \textit{learn} such an ordering. For each hidden layer to compress, we would like the following property to approximately hold: \[
    \forall k_1, k_2\in Z_+: k_1 < k_2 \implies g(h_{0\ldots k_1}(x)) \lesssim g(h_{0\ldots k_2}(x))
\]
Where $h_{0\ldots k}(x)$ are the first $k$ features of $h(x)$. This property can be thought of as being similar to integer encoding, where the higher bits of an integer encode more of its magnitude. In order to learn this property, we train the network using Structural Dropout (SD). Structural Dropout is defined as follows:

\begin{equation}\label{eq:sd_train}
    \text{SD}(x\in\mathbb{R}^N, p\in[0,1], \text{lb}\in\mathbb{Z}_+) = \begin{cases}
    x & \text{with probability $1-p$} \\
    \frac{N}{i} x_{0...i}\oplus\vec{0}, i\sim\mathbb{U}(\text{lb}, N) & \text{otherwise} \\
    \end{cases}
\end{equation}
Where $\text{lb}$ is some lower bound on the number of elements (defaulting to 1), $p$ is the dropout probability (defaulting to 0.5), and $i$ is an index chosen uniformly from all indeces, after which all elements are zeroed. There is an additional normalization constant $\frac{N}{i}$ so that $\mathbb{E}[x]$ is the same regardless of the number of sampled elements, and this normalization is applied both during training and inference. By randomly dropping out later indices, the network is forced to encode important information at lower indices, achieving the desired property. SD also retains some chance of training the entire state, otherwise the highest indices would be trained infrequently, and would not serve as supervision for the smaller layers, which we demonstrate later in ablations.

Structural Dropout acts as a method for training an entire ensemble of ``matryoshka'' networks, or networks that are nested within each other that share weights. During inference, we can pick from these nested networks to suit resource and accuracy constraints. Specifically, at inference time, we can make Structural Dropout deterministic and only compute:
\begin{equation}\label{eq:sd_test}
    \text{SD}_{\text{test}}(x, k\in[0,N]) = \frac{N}{k} x_{0...i} \oplus\vec{0}
\end{equation}
For some desired $k\in[0,N]$. Since the higher indices are \textit{all} $0$, it is easy to prune layers. For example, if we have
\[ A\in\mathbb{R}^{m,n}, B\in\mathbb{R}^{N,n}, x\in\mathbb{R}^N \]
\begin{equation*}
    A(SD_\text{test}(Bx, k)) = A[:, \text{:k}](\frac{N}{k}B[\text{:k},:]x)
\end{equation*}
Where $A[\text{:k},:]$ is Python slicing notation. We are able to prune linear layers applied before SD keeping $k$ rows, \textbf{and} prune linear layers after, keeping $k$ columns. Thus, we find that while the original matrix multiplications were $O(nN) + O(mn)$, we can reduce this to $O(kN) + O(mk)$. The concept of SD can be applied to many different kinds of layers, such as convolutional layers or self-attention layers, by dropping out later feature channels or attention heads, but we leave analysis of these to future work and focus only on expensive linear layers.

\begin{figure}[!ht]
    \centering
    \includegraphics[width=0.75\textwidth]{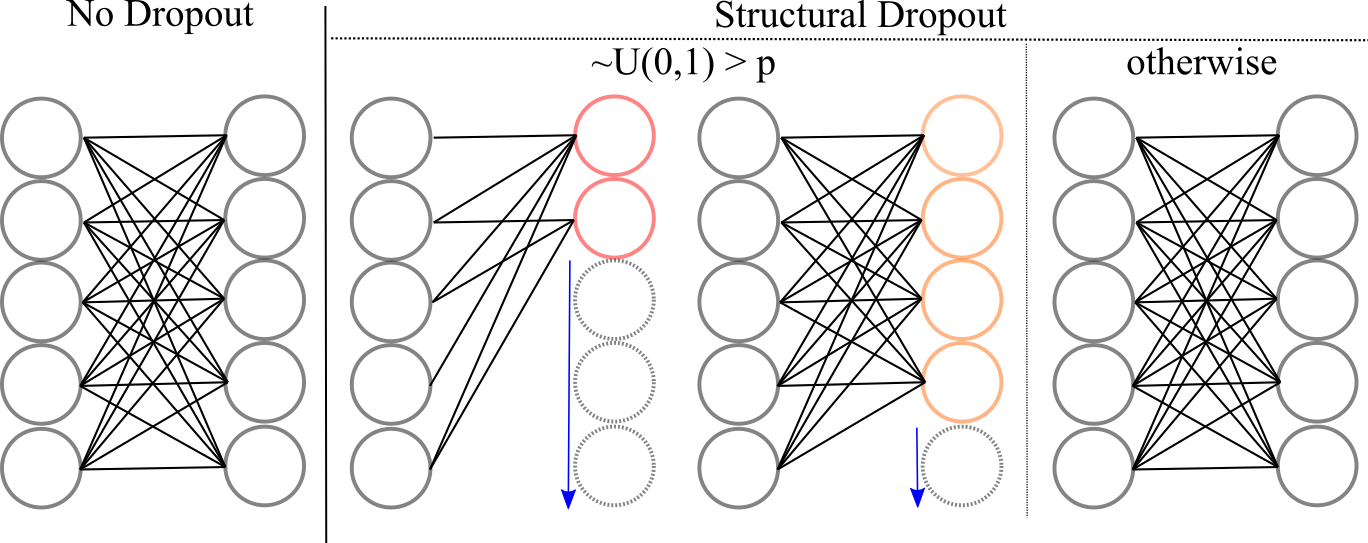}
    \caption{\textbf{Visualization of Structural Dropout}. Rather than randomly selecting indices to prune, during training Structural Dropout prunes all nodes after a uniformly randomly selected index, and normalizes the expectation based on the number of features dropped. With some probability, we run the full network, using it as supervision for the smaller networks indirectly.}
    \label{fig:sd}
\end{figure}

To summarize, this work's contributions are as follows:
\begin{enumerate}
\item A variant of Dropout, Structural Dropout, that trains an ensemble of nested networks, which can later be disentangled for compression without additional retraining.
\item A validation of Structural Dropout on 3 example tasks, demonstrating its effectiveness across a variety of methods while retaining accuracy.
\item An implementation of Structural Dropout, released at \url{https://github.com/JulianKnodt/structural_dropout}.
\end{enumerate}

%auto-ignore 
\section{Background}

Structural Dropout can be understood as model compression using knowledge distillation, but is heavily inspired by Dropout.

\subsection{Dropout}

Dropout~\cite{Dropout} was originally proposed as a methodology to prevent networks from overfitting by preventing neurons from strongly entangling information, preventing memorization of training data to specific outputs. This is done by replacing values in the hidden state with $0$ with probability $p$. The original authors offer one interpretation of this as an ensemble approach, where many smaller models are averaged together in order to achieve better generalization. Regardless of interpretation, Dropout prevents overfitting, and there have been a large number of variants of variants that extend Dropout to convolutional neural nets~\cite{SpatialDropout, devries2017cutout}, and other variations of dropout~\cite{DropConnect, FastDropout, VariationalDropout}. Our work is a variant of Dropout, but varies from prior work in that our goal is no longer generalization, but compression. Structural Dropout closely resembles \textit{Stochastic Depth}~\cite{StochasticDepth}, but instead of varying depth, ours can be considered an ensemble along model widths, and does not require additional skip connections.

\subsubsection{Targeted Dropout}

SD also resembles Targeted Dropout~\cite{TargetedDropout}, but instead of dropping independently from higher indices, as SD drops everything above a given index, and retains some probability for running the entire network. This leads to a few key differences, as Targeted Dropout will not be safe for pruning to arbitrary sizes, and it must be known ahead of time the target size to prune to. SD will also be an ensemble of $N-\textit{lower bound}$ models, where as Targeted Dropout will be an ensemble of $2^{N-k}$ models.

\subsection{Model Compression}

Model compression is important in the era of heterogenous computing, where ML models may have to run on mobile devices, older machines, or even inside of embedded micro-controllers. This conflicts with the continual increase of the size of state-of-the-art models, as many users are unable to immediately use them due to resource constraints.

For this work, we divide model compression into two forms: architectural (factorization, and knowledge-distillation), and non-architectural (pruning and quantization). We will mostly not focus on factorization, as it is not closely related to this work.

\subsubsection{Non-Architectural Compression}

``Non-architectural'' compression encompasses pruning and quantization, as they can be done without model architecture changes. Pruning is using regularization such as $\ell_1$ regularization in order to set many weights of a learned network to 0, reducing parameter count.

Quantization~\cite{NN8bit, DLLimitedNumericalPrecision} is the use of different representations such as integers or reduced bit floating point numbers to gain more efficiency at inference time. This has the benefit of greatly reducing memory, at the cost of reduced per operation accuracy. Some quantization, such as half-precision floating point, can be used during training.

Non-architectural compression is non-intrusive and can be used without modifying the training procedure while still providing a significant speed boost. The downside is that they are limited in the amount of savings for a given architecture. Quantization can only save a limited amount of memory, and ad-hoc pruning can only remove so much without a significant drop in accuracy.

Non-architectural methods do not modify the architecture of a neural net and are orthogonal to our approach, but we do not experiment with both.

\subsubsection{Architectural Compression}

Architectural compression compresses a neural net by modifying its architecture. For example, the Lottery Ticket Hypothesis~\cite{LotteryTicketHypothesis} finds small networks inside of a larger network to find embedded ``Lottery Ticket'' networks. Another methodology for architectural compression is the teacher-student model, where a large teacher network is used as a reference for a smaller student model. In Knowledge Distillation~\cite{Distilling}, a large network's output is used as a soft-target for a smaller network, achieving similar accuracy and compressing the network. Since the networks are trained independently, the smaller network's architecture may be different from the original network, or may just have smaller layers.

The benefit of architectural compression is that a network can be compressed beyond the original architecture. On the other hand, it requires deeper knowledge, such as selecting a new architecture for knowledge distillation, or introducing bias by factorizing layers. It also may require additional training, which for some users may be prohibitive.

\subsubsection{Once-For-All}

Once-For-All~\cite{cai2020once} is a similar approach to our work, performing compression on convolutional layers, looking at accuracy on ImageNet~\cite{ImageNet}. Once-for-all provides ``elastic'' width, depth and kernel-size, similar to SD which has dynamic width. The elastic width of convolutions is most similar to our work, but to determine which features are used in smaller networks, they rely on the $\ell_1$ norm of a channel's weight. Compared to Once-For-All, SD does not bake a heuristic into compression, and lets the network move the features around to best optimize the objective. Their approach is designed specifically for CNNs, whereas this work focuses on SD for linear layers, but can be dropped into any architecture. One other key difference of Once-For-All is that it requires fully training the whole network, before allowing for optimization of smaller networks. Instead of this, we stochastically sample the original network, which removes the additional stage of selecting when to prune and train smaller networks. Finally, SD trains all models jointly as opposed to fine-tuning smaller models.

\subsubsection{Slimmable Networks}

Another similar approach to Structural Dropout is Slimmable Networks~\cite{slimmableNN, uslimmableNN, autoslim}. Instead of using Dropout for dynamic channel width, Slimmable Networks uses batch normalization in order to be able to control the width of the network at inference time. It introduces an additional set of batch-normalization parameters for each target width, and a finite number of channel widths are selected. As compared to Structural Dropout, Slimmable Networks requires selecting a much smaller number of target widths, storing independent batch-norm parameters for each width, and modifying the training procedure to accumulate gradients across all chosen widths during each training iteration. Structural Dropout, on the other hand, does not require selecting target widths, adds no additional parameters, and is agnostic to the training procedure. As compared to Slimmable Networks, Structural Dropout acts as a drop-in layer into any architecture without any tuning, and is more flexible in that it will enumerate \textit{all} possible widths.

\textbf{Structural Dropout} trains a hierarchical ensemble of teacher-student networks that learn jointly, without requiring hyper-parameter selection. This helps achieve the non-intrusive ease of use that non-architectural methods use, while allowing compression of the model architecture.

As compared to prior work, Structural Dropout makes a few key modifications:
\begin{enumerate}
    \item SD removes the need to use some heuristic such as the norm of the weight in order to decide which layers are important, baking feature importance into the training procedure.
    \item SD requires no hyper-parameter search for target sizes, enumerating many smaller networks stochastically during training.
    \item SD can be dropped into an existing architecture, and does not require tuning or deep knowledge of the prior architecture, and is task-agnostic.
    \item SD does not require modifying the training procedure.
\end{enumerate}
%auto-ignore
\section{Method}

Structural Dropout is defined for training in Eq.~\ref{eq:sd_train} and inference in Eq.~\ref{eq:sd_test}. A diagram of Structural Dropout is shown in Fig.~\ref{fig:sd} We also provide a few more implementations and usage details, which are not explicitly outlined in the equations.

\subsection{Training Efficiency with Structural Dropout}

While training with Structural Dropout, we get the benefits of pruning that are achieved during inference. In this work, we add SD between linear layers, after activation functions. Since SD is applied after activation, even if an activation function is non-zero preserving ($\sigma(0) \neq 0$) the sparsity of the layer is retained, since the activation function cannot make the sparse vector dense. Because the vectors are guaranteed to be sparse, during training we can prune rows and columns from linear layers rather than explicitly passing $0$, since they are identical operations. This reduces the average cost of training the model.

\subsection{Coarse Structural Dropout}

Since Structural Dropout trains an ensemble of models, it may take longer to converge than the original model. On top of learning weights, the network learns to prioritize certain features over others, making the given task more difficult, albeit only before it has ordered features. In order to prevent a network from having to learn how to prioritize all features separately, it may be easier to learn clusters of features. For example, instead of selecting every value from 1 to $k$, SD takes a parameter to select every group of $n$ elements from 1 to $k$. This speeds up training since it reduces the number of models in the ensemble, at the cost of having fewer selections for the final model size.

\subsection{Lower Bounds on Features}

In many cases, it may be known to a practitioner that a certain number of features are necessary. For example, in a classification task over $n$ classes, theoretically at least $n-1$ features are necessary to produce a meaningful distribution over classes. By providing a lower bound on the cutoff index, we focus training time on strictly necessary features. We note that the network will only be slower to train if the cutoff is set too low, as the network will still attempt to optimize a small set of features, but the rest of the features will be unaffected since they were zeroed. Thus, lower bounding the feature size is an opt-in hyper-parameter, and can be left at a default of 1 without causing divergence.

\subsection{Selecting \textit{p}}

The choice of hyper-parameter $p$ affects how much Structural Dropout seeks to utilize the full model versus optimizing smaller models. Setting $p=0$, the smaller models are never optimized, and only the larger model is never trained, and is equivalent to not having dropout. Setting $p=1$, the full model is only trained with probability $\frac{1}{N}$, and the higher-indexed features get trained exponentially less. In expectation, $50$\% of the hidden features will be dropped out, which is contrast to standard Dropout which would have no elements. We opt for setting $p=0.5$ as a default. Within our experiments, we ablate setting $p=1$ to demonstrate that it is useful to optimize the entire model, as well as setting $p=0.01$ to demonstrate that it is necessary to have a higher $p$ to compress more.

\subsection{Inference Width Selection}

Because there are multiple layers with SD, it is computationally to expensive to evaluate all permutations of widths for the models. Thus, we arbitrarily choose to make all parameters equal: $w_1 = w_2 = \cdots = w_n$. This reduces the number of possible models from $w^n$ to $w$, and we exhaustively enumerate this set of models. We leave a better search for optimal widths to future work.
%auto-ignore
\section{Experiments}

Since our method is a layer that can be put into many architectures, the focus of this work is gaining an intuition and demonstrating its efficacy. To do so, we demonstrate how it works on 3 different experiments: an intuition gaining toy example with MNIST~\cite{deng2012mnist}, a demonstrative example of how it may be effective in compressing arbitrary architectures with PointNet~\cite{PointNet} and ModelNet10~\cite{ModelNet10}, and a demonstration of how it may be used in the SOTA with Local Implicit Image Functions (LIIF)~\cite{chen2021learning} for SISR.

For all of our experiments, we run on a single Intel-i7 CPU, with one NVIDIA GTX-1060. In all diagrams showing number of parameters, we add a line from the minimum parameter count to the maximum as a reference, since many are almost linear. All experiments are implemented in PyTorch~\cite{pytorch}.

\subsection{MNIST}

\begin{figure}
    \centering
    \includegraphics[width=\textwidth]{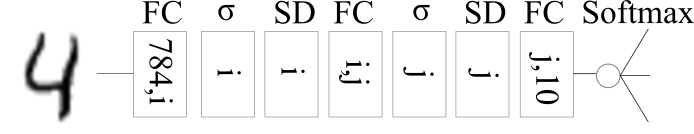}
    \caption{Architecture for MNIST with Structural Dropout. Structural Dropout allows for variable width fully-connected layers. During training, $i,j$ are independently and uniformly sampled from $[1,256]$, and at inference we set $i=j=c$, where $c$ is some constant.}
    \label{fig:mnist_arch}
    
    \begin{minipage}{0.45\textwidth}
    \centering
    \includegraphics[width=0.8\linewidth]{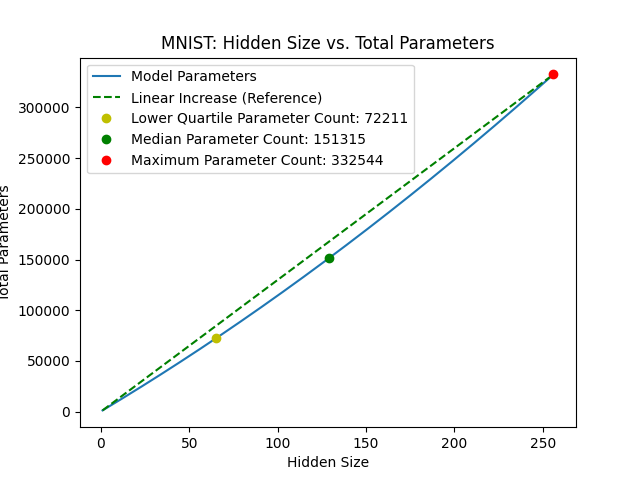}
    \caption{
    Size reduction of variable-width multi-layer perceptrons in MNIST. Since the hidden layer can vary both dimensions, reduction is more than linear.
    }
    \label{fig:mnist_param_count}
    \end{minipage}
    \hspace{5pt}
    \begin{minipage}{0.45\textwidth}
    \centering
    \includegraphics[width=0.8\linewidth]{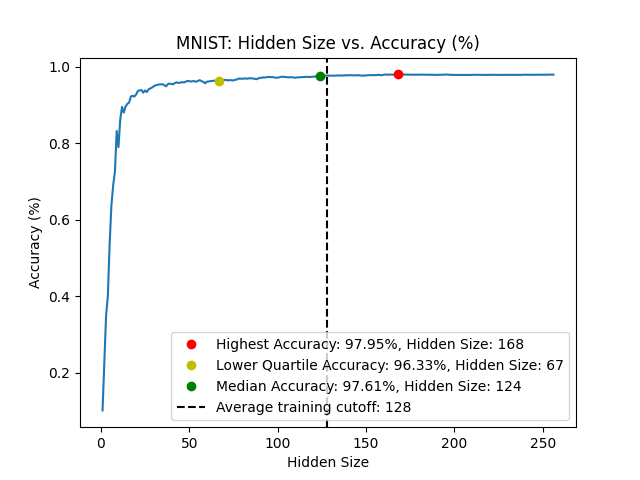}
    \caption{
    Structural Dropout is able to maintain accuracy at much smaller widths on MNIST in a single training procedure.
    }
    \label{fig:mnist_accuracy}
    \end{minipage}
    \label{fig:mnist}
\end{figure}

\begin{figure}[!ht]
    \centering
    
\end{figure}

To gain an intuition for using SD, we first try it on a toy example with MNIST~\cite{deng2012mnist}. Our architecture is a simple multi-layer perceptron with a hidden size of 256, shown in Fig.~\ref{fig:mnist_arch}. This architecture contains two Structural Dropouts, between two fully connected layers after their respective activation functions. The network is over-parametrized, each hidden layer has 256 neurons, and we seek to shrink it down as much as possible without significant accuracy degradation. We perform 100 epochs of training, using the Adam optimizer with a learning rate of $\num{8e-4}$, and standard cross-entropy loss.

After training, we then iterate over possible hidden sizes, $k$, and compute validation loss, to generate an accuracy versus resource trade-off comparison, as shown in Fig.~\ref{fig:mnist}. In addition, we also show a comparison of $k$ to the total number of elements, as an intuition for the size compression benefits.

With the full model, we see accuracy around 98\%, and with smaller sizes there is a steep increase in accuracy at around 15 parameters, from 10\%, equivalent to random chance, up to 80\%. Above that size, we see a much slower increase to 98\% accuracy on the test set. The optimal is not the model with all parameters, giving ``free'' compression: higher accuracy and reduced model size. We can also select a lower $k$, trading memory for accuracy. For example, we can select the lower quartile of accuracy at 96\%, with $\frac{1}{5}$ of the memory cost.

\subsection{PointNet}

\begin{figure}
    \centering
    \includegraphics[width=\textwidth]{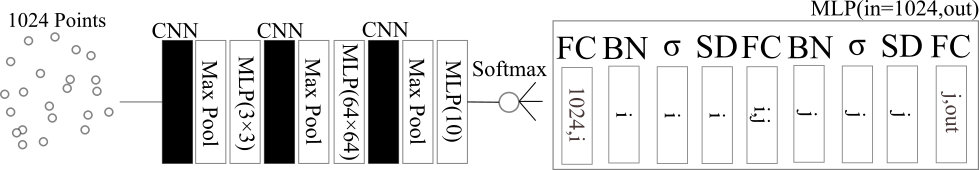}
    \caption{\textbf{Architecture for PointNet with Structural Dropout}. $\sigma$, is LeakyReLU, BN is batch normalization, and FC is fully-connected. We denote convolutional neural nets as black boxes, as we do not attempt to compress them in this work.}
    \label{fig:pointnet_arch}
\end{figure}

To demonstrate that our method can work on many architectures, we test SD on PointNet~\cite{PointNet}, a model for classifying point clouds. Since this task is not common, the model has not needed to be compressed, demonstrating that our method can work on arbitrary models without expensive grid search while using the original training procedure. In this work, we ignore the convolutional layers, and add additional SD layers where previously there were none. The architecture used for PointNet is shown in Fig.~\ref{fig:pointnet_arch}, where we add SD between linear layers. We experiment on ModelNet10~\cite{ModelNet10}, classifying 10 choices and optimizing with cross-entropy loss. Since batch normalization~\cite{BatchNormalization} is applied per feature in the original model, it is easily adapted to sparse layers. We train on ModelNet10 for 100 epochs, using the given train/test split.

\begin{figure}
    \centering
    \begin{minipage}{0.2\textwidth}
    \includegraphics[width=\linewidth]{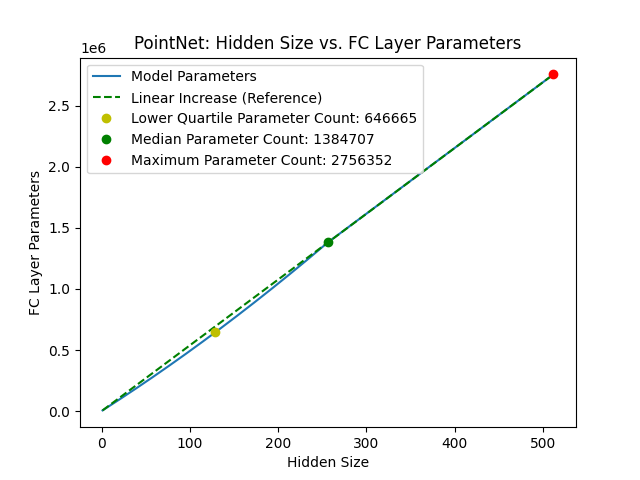}
    \caption{PointNet parameter savings are near linear.}
    \label{fig:pointnet_param_count}
    \end{minipage}
    \hspace{5pt}
    \begin{minipage}{0.75\textwidth}
    \includegraphics[width=0.32\linewidth]{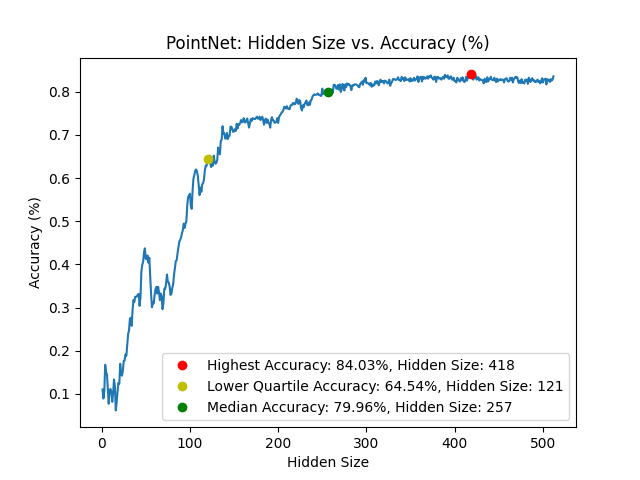}
    \includegraphics[width=0.32\linewidth]{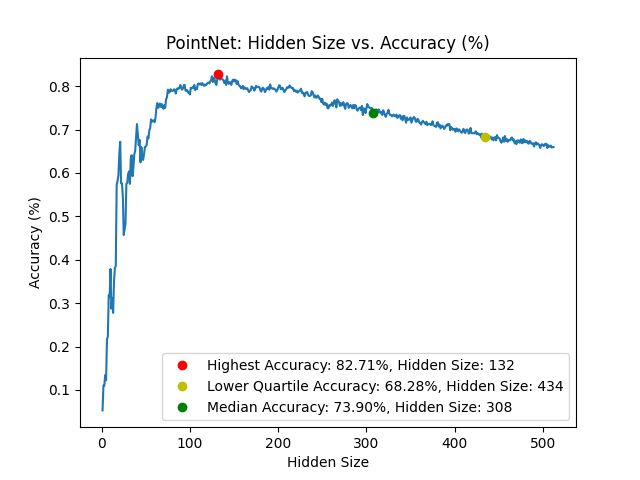}
    \includegraphics[width=0.32\linewidth]{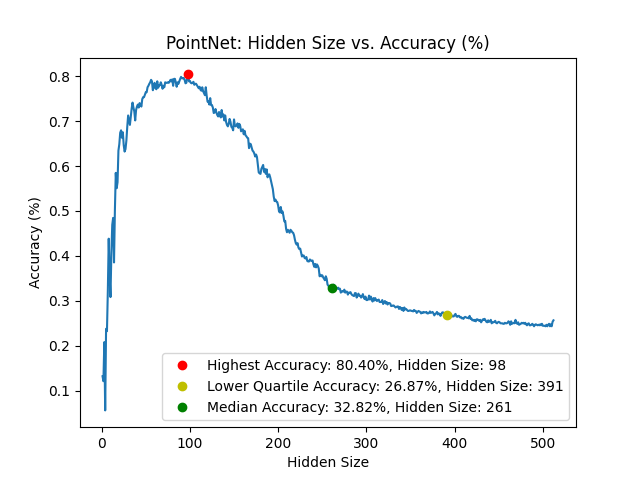}
    \caption{
        From left to right: $p=0.01,0.5,1$.
        Ablation of changing $p$ using PointNet~\cite{PointNet}. We find that Structural Dropout achieves better maximum accuracy with lower $p = 0.01$, and higher compression with higher $p = 1$.
    }
    
    \label{fig:pointnet_accuracy}
    \end{minipage}
    \caption{
        Within PointNet, we find that the highest validation accuracy is achieved with a much smaller number of parameters. We are thus able to reduce parameters by $\frac{1}{2}\text{ to }\frac{3}{4}$, \textit{and} improve performance.
    }
    \label{fig:pointnet-experiment}
\end{figure}

Results from a run of the experiment are shown in Fig.~\ref{fig:pointnet-experiment}. Surprisingly, we find that we are able to get performance improvements with a much smaller parameter size than the maximum budget, with 75\% of the original parameters without accuracy loss. We consider two possibilities for why. First, it may be the case that by sampling more models, it's more likely to find a better one among the set of possibilities, which is agnostic to Structural Dropout. Second, it could be that with online knowledge distillation smaller models learn important features from larger teacher models.

We also perform ablation of the parameter $p$ on PointNet. We observe that when $p=1$, there is a significant boost in compression without accuracy loss compared to the original model, but the accuracy of larger models is significantly degraded. When $p=0.5$, there is an accuracy boost, but requires more parameters than $p=1$. When $p=0.01$, there is even more of an accuracy boost, but at the cost of little compression. Specifically, we see better compression at $p=1$ at the same accuracy as the median value for $p=0.01$. Because we are able to achieve such high compression, it indicates that PointNet was significantly over-parameterized in its linear layers. If PointNet were to have to required all of its features, setting $p$ lower ensures that the full set of features gets trained.

\subsection{LIIF}

Finally, we demonstrate the effectiveness of Structural Dropout on a recent SOTA approach to single-image super resolution, LIIF~\cite{chen2021learning}. LIIF consists of an encoder and decoder, where the encoder is another super resolution architecture such as EDSR~\cite{EDSR}, and the decoder grid-samples some hidden state of EDSR, and passs it through a multi-layer perceptron to recover a high-resolution version of the original image. This approach is based off of recent advances in neural rendering that use an MLP to output feature vectors from 2D or 3D coordinates. One downside of using MLPs for low-dimensional inputs to other low-dimensional outputs is that they are very expensive, especially when they need to be queried at many different points. This motivates the need for compressing them, since they are queried many times and are the bottleneck of some pipelines for compute and memory.

To demonstrate SD for this use case, we modify LIIF in a few ways. First, between each linear layer in the MLP, we add a SD layer, with a lower bound of 32 and $p=0.5$. The lower-bound is a conservative estimate on the minimum features necessary, and $p=0.5$ is the default setting. We also find in experimentation that adding skip connections to the original architecture improves the convergence rate. Therefore, we add new sparse skip connections to our network to increase convergence speed. We also choose to include them to demonstrate how to adapt non-sparse structures to our approach, as many MLPs may use them.

In order to have sparse skip connections, we interleave two vectors rather than concatenate one after the other. Compared to normal stacking, if there are $n$ vectors which are sparse and the least sparse vector has $k$ trailing zeros, their interleaving will contain $nk$ trailing zeros.

Other than these modifications, we use the original architecture and train on Div2K~\cite{div2k} at $\frac{1}{4}$ resolution, with the Adam~\cite{adam} optimizer and learning rate $ = \num{3e-4}$.

\begin{figure}
    \centering
    \includegraphics[width=0.32\linewidth]{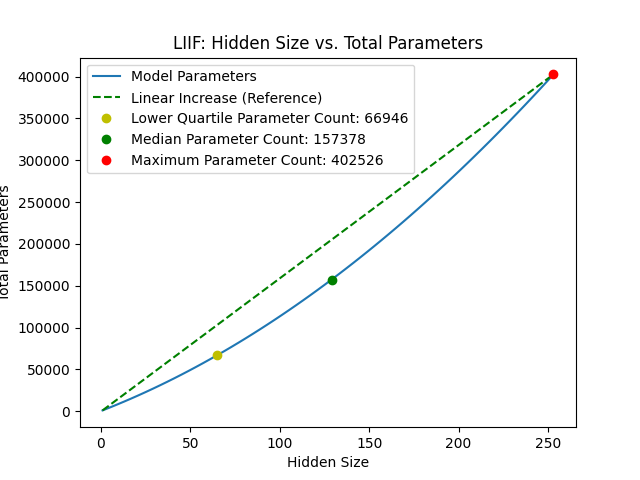}
    \includegraphics[width=0.32\linewidth]{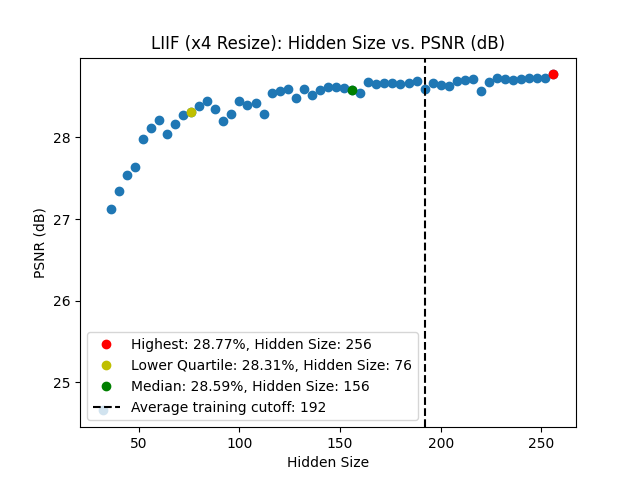}
    \caption{
        Using Structural Dropout with LIIF, we remain within 1 PSNR of accuracy at 80\% of the cost. We visualize every $4$th hidden size, as it is costly to validate all possible widths.
    }
    \label{fig:liif_acc}
\end{figure}

The original reported accuracy at $\times 4$ super resolution is 29 PSNR, and with SD achieves within $0.25$ PSNR. With SD, we are able to retain within 1 PSNR at 80\% compression, as shown in Fig.~\ref{fig:liif_acc}. We validate every 4th hidden size, due to the large cost of validating all hidden sizes. In a deeper network with skip connections and originally without Dropout, we are still able to add in Structural Dropout and compress the model. SD thus works as efficient compression as a drop-in layer or replacement for Dropout in existing architectures.

We note that in training LIIF, it became more difficult to measure performance, as sampling different model sizes lead to random accuracy jumps.
%auto-ignore
\section{Discussion}

Structural Dropout serves as a minimal addition to existing architectures to perform hyper-parameter search and compression. Since it does not require expensive retraining or additional domain knowledge, it is more easy to adopt than domain specific modifications, and is orthogonal to pruning and quantization.

Out of our experiments, it is clear that there is a steep cliff where information saturates, and that a significant features, between $50-80\%$, of features can be pruned with minimal accuracy change. When more aggressively pruning using a higher dropout rate, up to $80$\% can be pruned with no accuracy loss, as seen on PointNet.

Structural Dropout also may help performance, as purely by sampling multiple models, one of them may perform better on the given task.

\section{Limitations}

While our method is straightforward to add into existing architectures, there are some downsides to it.

One significant downside is that SD increases the search space, making the problem more difficult. Since the problem is more difficult, the training session may take longer, despite the speed up of using fewer parameters. This training time is not significantly longer than the original, but it is not clear how much longer. On top of increased training time, comparing all channel widths for validation loss is slow, since there are a large number of models to test. This can be parallelized easily if resources are available, as unlike training the model will be read-only, otherwise a sparse search can be performed.

Another downside of SD is that it becomes more difficult to validate and track convergence during training. Since the model may randomly have lower accuracy during training due to low channel width, a model's convergence becomes difficult to determine. Thus, it makes sense to use SD for models which have been trained before, and prior convergence parameters have already been set. On these models, it can also be used as a way to perform hyper-parameter search for channel width.

In addition, while we assume that all the SD sizes should be the same at inference, in a larger model with a variety of layers this may not be the case. To perform an exhaustive search over all possible choices of channel size would be expensive, so an efficient search strategy is important. One common assumption is that each choice can be made independently, but we leave exploration of this to future work.

Finally, our method is unable to prune entire layers, as SD is limited to changing the width of a neural network but cannot modify its depth. Thus, a network where the performance bottleneck is depth will not be able to get as much benefit.

\section{Future Work}

While we show Structural Dropout's application on FC layers, the same principle can be extended to other kinds of layers. For example, the same idea is applicable along the channel dimension of CNNs, allowing for pruning of features. Another possible extension is for transformers, where it is important to select a number of attention heads. Structural Dropout could be applied to both attention head size, and number of attention heads. We hope that this work is extended to explore use of Structural Dropout in a broad variety of architectures for practical and efficient compression.
%auto-ignore
\section{Conclusion}

Structural Dropout is a method for inference-time compression, and can be used as a drop-in layer to existing architectures to achieve substantial compression, with minimal accuracy loss. It is the first methodology that the authors are aware of that allows for training many arbitrarily compressed models in a single training session, and then being to deploy them all jointly, at the cost of only deploying the largest model. From our experiments, we hope that it is evident that Structural Dropout is a powerful tool for compressing neural networks.

\begin{ack}

No funding was received for this work.

\end{ack}

\bibliographystyle{plain}
\bibliography{ref}

\end{document}